\documentclass{article}
\usepackage{ijcai26}

\usepackage{times}
\usepackage{soul}
\usepackage{url}
\usepackage[hidelinks]{hyperref}
\usepackage[utf8]{inputenc}
\usepackage[small]{caption}
\usepackage{graphicx}
\usepackage{amsmath}
\usepackage{amsthm}
\usepackage{booktabs}
\usepackage{algorithm}
\usepackage{algorithmic}
\usepackage[switch]{lineno}
\usepackage{amssymb}

\title{\texorpdfstring
{Learning From Failures: Efficient Reinforcement Learning Control\\
with Episodic Memory}
{Learning From Failures: Efficient Reinforcement Learning Control with Episodic Memory}}

\author{
    Chenyang Miao
    \affiliations
    Xi'an Jiaotong University
    \emails
    cy.miao@stu.xjtu.edu.cn
}

\begin{document}

\maketitle

\begin{abstract}
Reinforcement learning has achieved remarkable success in robot learning. However, under challenging exploration and contact-rich dynamics, early-stage training is frequently dominated by premature terminations such as collisions and falls. As a result, learning is overwhelmed by short-horizon, low-return trajectories, which hinder convergence and limit long-horizon exploration. To alleviate this issue, we propose a technique called Failure Episodic Memory Alert (FEMA). FEMA explicitly stores short-horizon failure experiences through an episodic memory module. During interactions, it retrieves similar failure experiences and prevents the robot from recurrently relapsing into unstable states, guiding the policy toward long-horizon trajectories with greater long-term value. FEMA can be combined easily with model-free reinforcement learning algorithms, and yields a substantial sample-efficiency improvement of 33.11\% on MuJoCo tasks across several classical RL algorithms. Furthermore, integrating FEMA into a parallelized PPO training pipeline demonstrates its effectiveness on a real-world bipedal robot task.
% Furthermore, the PPO parallel training framework integrated with FEMA demonstrates strong learning ability in the real bipedal robot task. 

% This leads to an abundance of short-horizon, low-return trajectories, ultimately hindering convergence and long-horizon exploration.
% resulting in a prevalence of short-horizon, low-return trajectories that hinder convergence and limit long-horizon exploration. 
% several classical reinforcement learning algorithms integrated with FEMA have achieved a substantial sample efficiency improvement of 33.11\% in the MuJoCo tasks. 
\end{abstract}

\section{Introduction}
Reinforcement Learning (RL) enables agents to progressively acquire complex skills through trial-and-error interactions. In recent years, RL has achieved significant success in the field of robot learning, demonstrating strong performance in locomotion~\cite{hwangbo2019learning,nahrendra2023dreamwaq,song2024learning} and robotic manipulation~\cite{wang2019learning,luo2025precise,lai2025roboballet} tasks. Despite these advances, deploying RL for robot control remains challenging: under high-dimensional continuous action spaces and tightly coupled contact dynamics, early-stage training is often dominated by premature terminations, such as collisions and falls, resulting in a large number of short-horizon, low-return trajectories~\cite{haarnoja2018learning,liu2022safe,tang2025deep}. The difficulty of obtaining long-horizon, high-quality training data constitutes a key bottleneck behind the notorious sample inefficiency of robot learning tasks.
% 难以获取长序、高质量数据是造成robot learning任务样本效率极其低下的重要原因。
% 智能体在训练前期无法探索得到长序列、高质量的动作序列，极大影响学习性能，带来严重的样本效率问题。
% As these low-quality experiences overwhelm the replay distribution, they obscure informative data, restrict effective exploration, and ultimately slow learning, thereby exacerbating to the notorious sample inefficiency of RL in robotic control. 

\begin{figure}[h]
\centering
\includegraphics[width=0.8\columnwidth]{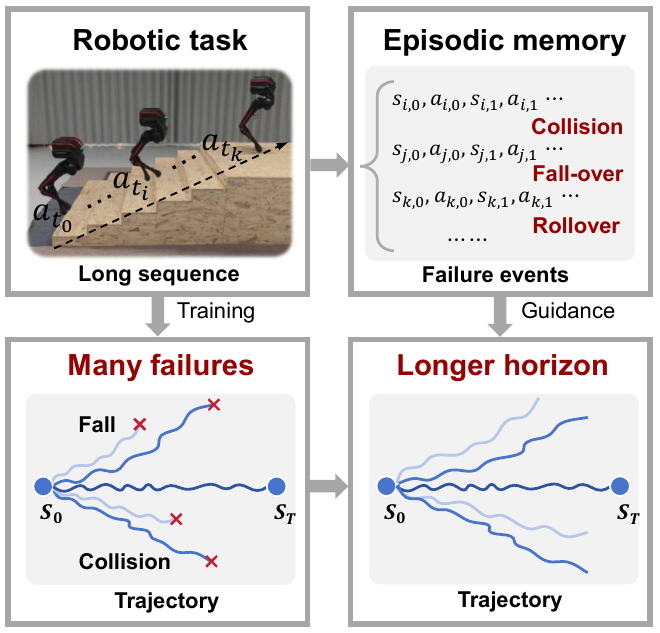}
\caption{Robotic agents frequently experience premature terminations during training. FEMA retrieves past similar failure experiences from the episodic memory to guide the robot away from previously experienced hazardous states, promoting longer-horizon exploration.}
\label{figure:cover}
\end{figure}

However, these short-horizon, low-return failure trajectories are informative rather than useless~\cite{thananjeyan2021recovery}.
Their implicit value does not lie in individual transitions, but in the spatiotemporal structure of the whole trajectory: how the robot moves through a sequence of states under a sequence of actions, and how certain state–action patterns consistently correlate with subsequent sudden termination.
Changes in state–action sequences can reveal and even anticipate the system’s transition toward hazardous configurations, providing early warning signals for imminent collisions or falls~\cite{xie2022ask}. 
Yet, transition-wise sampling breaks temporal dependencies within an episode and weakens the association between risk-indicative patterns and subsequent termination. As a result, failure experience is underutilized and can even negatively bias learning due to its prevalence in early training.
Inspired by human cognitive mechanism, episodic control constructs an external episodic memory to store and retrieve past experiences as events, which are leveraged to improve value estimation or guide decision-making~\cite{lengyel2007hippocampal,blundell2016model,lin2018episodic}. 
Prior studies have shown that episodic memory can significantly improve sample efficiency in discrete control tasks~\cite{lin2018episodic}, and has been further extended to continuous-action robotic control scenarios~\cite{zhang2019asynchronous,kuznetsov2021solving,liu2024bi}. Recent works realize efficient retrieval through learned task-aware embeddings for semantic matching~\cite{na2024efficient}. Nevertheless, existing approaches are predominantly success-centric, relying on high-return trajectories. 

Can we construct a failure-centric episodic memory module that leverages the abundant short-horizon failure trajectories generated during early training?

To mitigate the learning inefficiency caused by the prevalence of short-horizon, low-value trajectories during early training phase in the robotic task, we propose a novel technique: Failure Episodic Memory Alert (FEMA). 
As illustrated in Figure~\ref{figure:cover}, by recalling failure trajectories, FEMA discourages the robot from repeatedly relapsing into unstable states and promotes more effective exploration, thereby improving learning performance.
% As illustrated in Figure~\ref{figure:cover}, through the recall of failure trajectories, FEMA prevents the robot from continually relapsing into unstable states, encourages it to explore backwards, collects long-horizon, high-quality data, and finally improves learning performance. 
%有逻辑性质，FEMA的核心（收集数据，语义embedding；收集得到fem之后，我们提出一个新机制，怎么起作用）
Specifically, FEMA aggregates prematurely terminated trajectories from the early training phases of robot control tasks. Learning-based joint state–action embeddings are employed to characterize hazardous state–action pairs, enabling efficient memory retrieval and guidance.
% We utilize learning-based joint state-action embeddings to characterize hazardous state-action pairs for efficient memory retrieval and guidance. 
% 我们引入了 a risk-aware action selection module，让构建完成的failure episodic memory直接介入机器人的输出动作选择。通过在记忆中搜寻和当前状态相近的失败记忆，相关失败事件会被取出并用于指导机器人的实时动作输出。提出的风险评分机制指导robotic agent在多个候选动作中进行选择，以避免机器人再陷入过去相似危险状态，使其充分向后探索高质量数据加速训练进程。
We further introduce a risk-aware action selection module that enables the constructed failure episodic memory to directly intervene in the robot’s interaction with the environment. Failure events relevant to the current state are retrieved and leveraged to guide online action selection. By scoring candidate actions based on their associated risk, the proposed scheme discourages the robot from revisiting previously observed hazardous regions, thereby improving the quality of collected experience and accelerating training.
% The proposed risk-scoring scheme chooses among candidate actions to prevent relapses into previously observed hazardous regions, thereby improving the quality of collected experience and accelerating the training process.
% Building on this representation, we introduce a risk-aware action selection module that incorporates a risk assessment mechanism. By comparing the current state–action pair against stored failure experiences to estimate its associated risk, this module actively intervenes in action selection during exploration. 
% By retrieving the similar failure memories to the current state,
Empirical results substantiate that iteratively revisiting these ``painful lessons" effectively accelerates learning. Integrating FEMA significantly bolsters early-stage training stability and yields substantial improvements in sample efficiency. 

% We introduce FEMA, a failure-centric episodic memory framework，聚焦robot learning任务前期大量短序失败轨迹的重复利用，以提升算法在机器人任务中的表现。
% We propose Failure Episodic Memory Alert (FEMA), a novel framework tailored for robotic control tasks. By capitalizing on the abundant short-horizon, low-value trajectories often discarded during early training, FEMA effectively enhances both sample efficiency and learning stability.
%MuJoCo benchmarks上大量的实验结果证明了
% FEMA is a plug-in module compatible with multiple model-free RL algorithms, and extensive experiments on MuJoCo benchmarks show its 
The main contributions of this paper are summarized as follows:
\begin{itemize}
    \item We introduce FEMA, a failure-centric episodic memory technique that stores failure trajectories and leverages them for guiding the robot's more effective exploration in robotic control tasks.
    \item FEMA is a plug-in module compatible with multiple model-free RL algorithms. Extensive empirical evaluations on MuJoCo benchmarks demonstrate that our approach yields substantial improvements in sample efficiency and final performance compared to standard baselines.
    \item We further validate FEMA in a real-world bipedal robot stair-climbing task by incorporating it into a parallelized PPO training pipeline, highlighting its practical potential for real robot systems.

\end{itemize}

\section{Related Work}

%这个子章节，主要讲一下model-free强化学习算法的演进，以及解释replay buffer和优先级采样；再抨击没有充分利用不同state action之间的关联性；
\subsection{DRL and Experience Reuse}
% Classic deep reinforcement learning algorithms include value-based methods such as DQN~\cite{mnih2013playing}, actor–critic methods such as DDPG~\cite{lillicrap2015continuous}, TD3~\cite{fujimoto2018addressing}, and SAC~\cite{haarnoja2018soft}, as well as policy-based methods such as PPO~\cite{schulman2017proximal}.
Deep reinforcement learning has advanced through value-based methods such as DQN~\cite{mnih2013playing}, off-policy actor-critic methods including DDPG~\cite{lillicrap2015continuous}, TD3~\cite{fujimoto2018addressing}, and SAC~\cite{haarnoja2018soft}, as well as on-policy policy optimization methods such as PPO~\cite{schulman2017proximal}. 
% Classic deep reinforcement learning algorithms include value-based methods such as DQN~\cite{mnih2013playing}, DDPG~\cite{lillicrap2015continuous}, TD3~\cite{fujimoto2018addressing}, SAC~\cite{haarnoja2018soft} and policy-based methods such as PPO~\cite{schulman2017proximal}. 
However, these algorithms generally rely on large amounts of interaction data, making sample inefficiency a major bottleneck in complex control tasks. To alleviate this issue, experience replay was introduced by storing and repeatedly sampling past transitions to improve data efficiency~\cite{mnih2013playing}. Initially proposed with DQN algorithm, replay buffer has been widely adopted in DDPG, TD3, SAC, and more recent methods such as CrossQ~\cite{bhatt2024crossq}, becoming a core component of modern DRL methods. Prioritized Experience Replay (PER)~\cite{schaul2015prioritized} further improves replay efficiency by sampling transitions based on learning utility with importance sampling correction. Despite these advances, most replay-based methods still treat experiences as independent transition tuples and focus on reusing individual samples, without explicitly modeling event-level structural relationships among different experiences.

\subsection{Episodic Control}
% Episodic memory is a form of long-term explicit memory that encodes individual experiences together with their temporal and spatial contexts, and is widely regarded as a fundamental cognitive mechanism linking past experience to intelligent behavior.
Episodic memory is a form of long-term explicit memory that encodes experiences as events, explicitly preserving the temporal and spatial context in which those memories are generated~\cite{kasap2010towards,irish2013pivotal,moscovitch2016episodic}. 
Inspired by this mechanism, episodic control has been introduced into reinforcement learning as a means of improving sample efficiency by retrieving similar experiences from episodic memory module during learning or decision-making to accelerate policy convergence~\cite{blundell2016model,pritzel2017neural,lin2018episodic}.
% 现在应该讲他的局限性，从memory结构上说。
Efficient episodic memory retrieval remains a key bottleneck in episodic control. Random projection~\cite{dasgupta2003elementary} is commonly used to compress global states for fast retrieval, but it fails to preserve semantic similarity, forcing the use of small distance thresholds that limit effective reuse of goal-relevant experiences. Recently, NEC~\cite{pritzel2017neural} estimates action values via a differentiable neural dictionary and GEM~\cite{hu2021generalizable} develops state-action values of episodic memory in a generalizable manner. In contrast, EMU~\cite{na2024efficient} explicitly learns a task-relevant semantic embedding space for episodic retrieval, enabling more flexible and accurate matching of experiences that are semantically similar, thereby substantially improving the effectiveness of episodic memory utilization.

In robot learning, episodic control has been extended to both discrete and continuous action spaces. EMDQN~\cite{lin2018episodic} improves sample efficiency in discrete tasks by integrating episodic memory with value-based learning, while AE-DDPG~\cite{zhang2019asynchronous} and EMAC~\cite{kuznetsov2021solving} incorporate episodic control into deterministic policy gradient frameworks for continuous robotic control. Beyond guiding Q-value updates, BEMG~\cite{liu2024bi} directly leverages successful trajectories stored in episodic memory to guide action selection, enabling a more behavior-oriented exploitation of episodic experience. 
A notable limitation of existing episodic memory approaches is that, regardless of whether they are used for Q-value updates or action guidance, they prioritize high-value trajectories while overlooking the numerous short-horizon, low-value events prevalent during early training.
% A notable limitation is that existing episodic memory approaches is that, regardless of their use in Q-value updates or action guidance, they prioritize high-value trajectories while ignoring the numerous short-horizon, low-value events prevalent during early training.

%然而值得注意的是，不管是让情景记忆模块直接介入Q-value的更新，还是使用情景记忆进行动作指导，这里的episodic memory模块都聚焦于维护高价值的事件轨迹,往往忽略训练期间，尤其是训练中前期大量存在的短续、低价值事件。
% In control settings, episodic control has been successfully extended to both discrete and continuous action spaces. Lin et al. proposed EMDQN for discrete-action tasks, achieving significant gains in sample efficiency by integrating episodic memory with value-based learning. For continuous control, Zhang et al. and Kuznetsov et al. extended episodic control to deep deterministic policy gradients through AE-DDPG and EMAC, respectively, demonstrating the applicability of episodic memory mechanisms in robotic learning scenarios. Beyond using episodic memories to guide value-function updates by storing states and their maximum cumulative returns, Liu et al. further proposed BEMG, which directly exploits successful task trajectories stored in episodic memory to guide action selection at each decision step.

\section{Method}
% In this section, we propose a technique termed Failure Episodic Memory Alert (FEMA),尝试探索对short-horizon，low-return轨迹的reuse。FEMA包括两个部分：（1）Failure Episodic Memory Module；（2）Risk–aware action selection。FEMA可与任意model-free强化学习方法结合，在robot learning任务中可以有效prevents the robot from recurrently relapsing into unstable states。
%最后一句句式有点重复，可以再修改打磨一下。这里引出我们的框架图出来

\begin{figure*}[h]
\centering
\includegraphics[width=2.0\columnwidth]{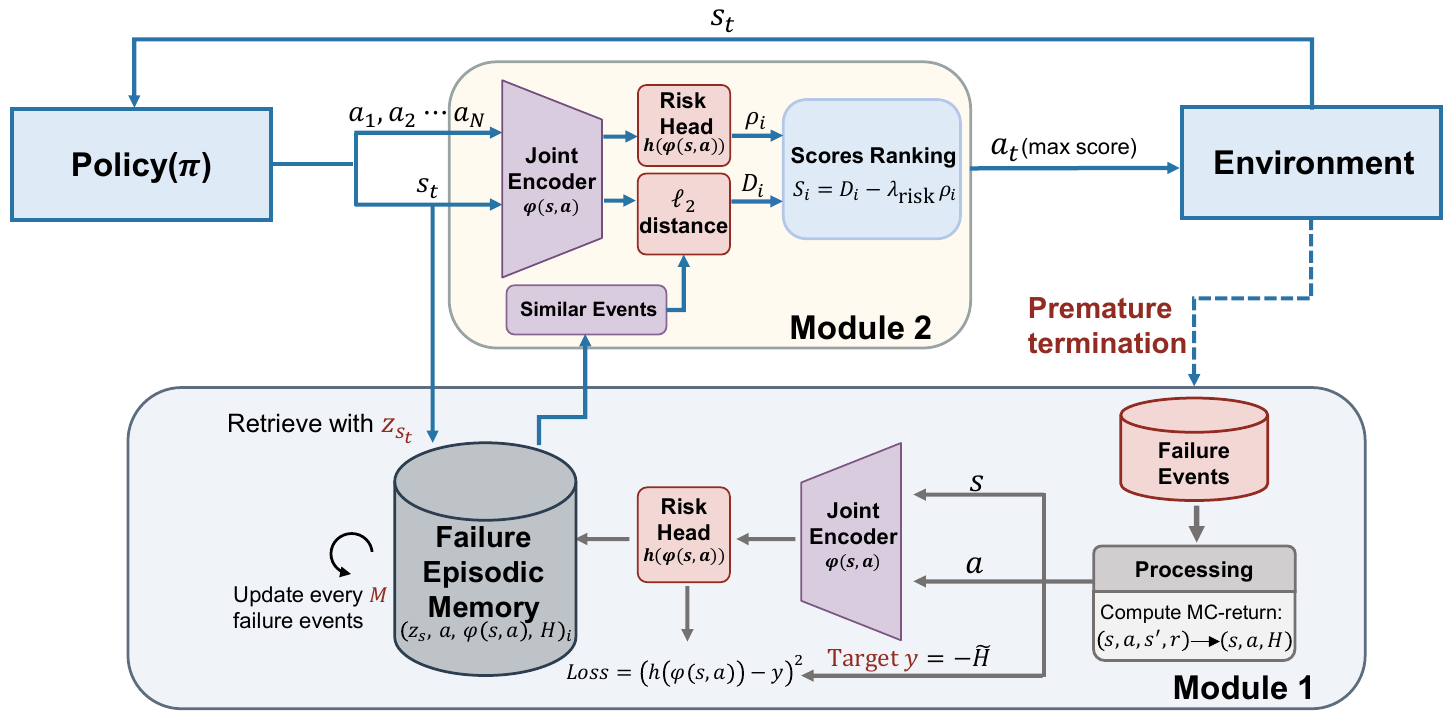}
\caption{Framework of Failure Episodic Memory Alert (FEMA)}
\label{figure:alg_framework}
\end{figure*}

\subsection{Overview}
In this section, we propose a technique termed Failure Episodic Memory Alert (FEMA), which aims to explore the reuse of short-horizon, low-return trajectories combined with episodic memory. As illustrated in Figure~\ref{figure:alg_framework}, FEMA consists of two key components: (1) Failure Episodic Memory Construction; (2) Risk-aware Action Selection Mechanism. FEMA can be integrated with diverse model-free reinforcement learning methods.
It effectively prevents the robot from recurrently relapsing into unstable states and encourages the exploration of long-horizon, high-quality experiences, thereby improving performance on robotic tasks.

\subsection{Failure Episodic Memory Construction}

% 1.确定失败事件，从后往前收集轨迹；
% 2.将(s,a,s',r)收集为(s,a,H);
% 3.joint state-action embedding的处理，以及rish head的设置，如何进行监督训练；
% Robot learning任务中，当机器人发生碰撞、falls等意外情况，导致当前训练回合非正常中止，我们将这个回合称作失败回合。我们从该回合的最后一步，即机器人训练回合意外中止的最后一步，我们将从最后一个时间步向前回推的k个时间步的轨迹称为失败情景记忆。我们按照事件的形式去存储每个失败回合对应的失败记忆，每一个失败记忆都由一串连续事件步的状态转移元组(s,a,s.r')组成。
As shown in the lower part of Figure~\ref{figure:alg_framework}, the failure episodic memory construction module is responsible for collecting, processing hazardous experiences arising during training. In robot learning tasks, an episode is defined as a failure episode when the training rollout terminates accidentally due to unexpected situations such as collisions or falls. For each failure episode, we record and store the state transition tuples $(s_t,a,s_{t+1},r)$ from the last $K$ time steps of the trajectory, encapsulating them as failure events. 

\begin{figure}[h]
\centering
\includegraphics[width=0.6\columnwidth]{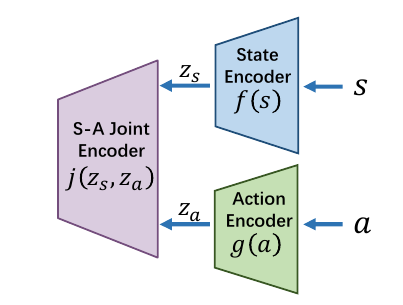}
\caption{Joint state-action encoder $\varphi(s,a)$}
\label{figure:encoder_framework}
\end{figure}

We process each failure event by computing the Monte-Carlo return $H$ associated with the state component of each transition tuple:
\begin{equation}
H_t \;=\; \sum_{n=t}^{T-1} \gamma^{\,n-t}\, r_n ,
\label{eq:return_to_go}
\end{equation}
where $T$ represents the actual termination time step of this failure event.
% \begin{equation}
% H=r_{t}+\gamma r_{t+1}+\cdots+\gamma^{k-1}r_{t+k-1}
% %写一个累计奖励H的公式
% \end{equation}

Inspired by the efficient memory embedding proposed in EMU~\cite{na2024efficient}, we propose a learning-based joint state-action embedding framework, trying to construct the semantic meaning of state-action pairs. 
%这里放一张joint state-action embedding的解构图
% Specifically, we first introduce a state encoder $f{(s)}$ and an action encoder $g{(a)}$ to address the significant discrepancy between the state space and the action space.
% 在FEMA的方法框架图中，我使用了$\varphi(s,a)$来代表joint state-action embedding，他的详细框架图如图2所示。
In the FEMA framework diagram, we use $\varphi(s,a)$ to denote the joint state–action embedding, and its detailed architecture is shown in Figure~\ref{figure:encoder_framework}.
To bridge the discrepancy between the high-dimensional state space and the continuous action space, we introduce a pair of encoders $(f, g)$ to learn compact and structured representations.
Specifically, the state encoder $f(s)$ maps the raw state $s$ into a latent state embedding $z_s$, while the action encoder $g(a)$ maps the action $a$ into an action embedding $z_a$:
% 放state encoder和action encoder的公式；
\begin{equation}
z_s := f(s), \quad
z_a := g(a).
\label{eq:state_action_encoder}
\end{equation}

The joint state–action embedding $j(z_s, z_a)$ takes the state embedding and the action embedding as inputs and encodes them into a unified representation. 
For notational simplicity, we denote the joint state–action embedding as $\varphi(s,a) = j(z_s, z_a)$.

% Leveraging the information captured by the joint embedding, we design an additional risk head $h(\cdot)$ to estimate the hazardousness of the current state-action pair. 
% % Leveraging the rich information captured by the joint state–action embedding, we further design a risk head. 
% We use $-H$ as the supervisory signal, where higher returns correspond to lower risk, 
% and thus $-H$ serves as a natural risk surrogate. 
% Specifically, we optimize the mean-squared error objective over batches sampling from the failure episodic memory $\mathcal{D}$:
% % 我们在方法框架图中使用$\varphi(s,a)$来概括了整个joint state-action embedding.
% \begin{equation}
% \mathcal{L} = \mathbb{E}_{(s,a,y)\sim\mathcal{D}}\!\left[\left(h(\varphi(s,a)) - y\right)^2\right],y=-H
% \end{equation}

Leveraging the information captured by $\varphi(s,a)$, we design an additional risk head $h(\cdot)$ to estimate the hazardousness of the current state--action pair.
We use the negative Monte Carlo return $-H$ as supervision, where lower returns indicate higher risk. 
To stabilize training and mitigate scale variations of returns across batches, we apply a $z$-score normalization to the Monte Carlo returns within each training batch:
\begin{equation}
\tilde{H} = \frac{H - \mu_H}{\sigma_H + \epsilon},
\label{eq:H_norm}
\end{equation}
where $\mu_H$ and $\sigma_H$ are the mean and standard deviation of $\{H\}$ in the batch, and $\epsilon$ is a small constant like $10^{-6}$ for numerical stability.
The final training target is then defined as $y = -\tilde{H}$.

Specifically, we optimize the mean-squared error objective over batches sampled from the failure episodic memory $\mathcal{D}$:
\begin{equation}
\mathcal{L} = \mathbb{E}_{(s,a)\sim\mathcal{D}}
\left[\left(h(\varphi(s,a)) - y\right)^2\right],
\quad y = -\tilde{H}.
\label{eq:risk_head_loss}
\end{equation}

% where $\mathcal{D}$ denotes the failure episodic memory.

% 放一个MSE的公式

% By training the joint embedding $\phi(s,a)$ and the risk head $h(\cdot)$ end-to-end under this supervision, the learned latent representation becomes more aligned with failure-related patterns. On the one hand, the semantic state embeddings enable efficient retrieval from the failure episodic memory compared to the general random projection methods. On the other hand, 
% 更重要的，为后面的评分机制服务
% learning-based joint state-action embedding的引入为后面候选动作
% And more importantly, nearest-neighbor retrieval in the embedding space is better matched to historically hazardous experiences, while the risk head provides an estimate of the risk associated with the current state--action pair.

% By using the Monte Carlo return as the supervision signal, the joint embeddings of hazardous state–action pairs are pulled closer in the latent space, and the risk head estimates the risk associated with the current state–action pair. 

By using the normalized Monte Carlo return $-\tilde{H}$ as the supervision target, we optimize the risk regression objective end-to-end. The loss is back-propagated not only to the risk head $h(\cdot)$ but also through the joint embedding $\varphi(s,a)$. 
Consequently, hazardous state–action pairs yielding low returns are encoded into a coherent risk-aware latent space, while the risk head estimates the risk of the current state–action pair.
Apart from that, the semantic state embeddings $z_s$ enable efficient retrieval from the failure episodic memory. The final failure episodic memory is organized as $\left( z_{s_i},\, a_i,\, \varphi(s_i,a_i),\, H_i \right)$. 

% By using the Monte Carlo return as the supervisory signal and updating the whole network through backpropagation, the joint embeddings of hazardous state–action pairs are pulled closer in the latent space, while the risk head estimates the risk associated with the current state–action pair. 
% Apart from that, the semantic state embeddings enable efficient retrieval from the failure episodic memory. The final failure episodic memory is organized as $\left( z_{(s_i)},\, a_i,\, \varphi(s_i,a_i),\, H_i \right)$. 

Instead of being updated at every step, the failure episodic memory is updated periodically by aggregating $M$ newly collected failure events.

% 这里Failure episodic memory采用定期更新的方式，每收集得到M个failure events进行一次集中更新。

%这里EC更新的方式还没说，这里我们采用定期更新，每收集得到多少个失败event就几种更新一次。

%以Monte-Carlo return为监督信号，使得“危险的” state-action pairs对应的joint embedding能够相互聚集,同时risk head能够输出当前state action的危险程度。最终的Failure episodic memory的存储信息格式为：

%翻译

%我们处理每个失败事件，计算每个event中的每个状态转移元组中的state对应的累计奖励H。
%受EMC工作的启发，我们尝试去构建joint state-action embedding，尝试构建包含语义信息的embedding。具体来说，我们首先设置了state-encoder和action-encoder，来解决状态空间和动作空间差异过大的问题。Joint state-action embedding接收state embedding和action embedding的输入，编码得到joint embedding。我们利用joint state-action embedding包含的丰富信息，设置risk head，使用-H作为监督信息。

%在-H的监督下，joint embedding包含语义信息，危险的embedding聚集，同时risk head能够很好反映出当前状态-动作组合的风险程度。

\subsection{Risk-aware Action Selection Mechanism}\label{s3-3}
% 1.候选动作组合（从分布中采样）
% 2.使用包含语义信息的state embedding查找，比较L2 distance;
% 3.根据H选择Top-K个动作，然后计算joint embedding的L2distance,综合risk head的输出综合打分；
The risk-aware action selection mechanism is illustrated in the upper part of Fig.~\ref{figure:alg_framework}.
Following the establishment of the failure episodic memory, we incorporate it into the learning loop through a risk-aware action selection mechanism. At each step, the agent acquires the current state and samples from a noisy distribution to generate a candidate action set comprising $N$ actions:
%在这里加一个带噪声的分布中采样的公式
\begin{equation}
a_i \sim \pi_\theta(\cdot \mid s_t), \quad i = 1, \dots, N ,
\label{eq:candidate_action_sampling}
\end{equation}
where $\pi_\theta(\cdot \mid s_t)$ denotes a stochastic policy parameterized by $\theta$, and $N$ is the number of sampled candidate actions.

For PPO-based and SAC-based algorithms in continuous control, the stochastic policy is commonly parameterized as a diagonal Gaussian distribution:
\begin{equation}
\pi_\theta(a \mid s) = \mathcal{N}\!\big(\mu_\theta(s),\, \mathrm{diag}(\sigma_\theta^2(s))\big),
\label{eq:stochastic_policy}
\end{equation}
where $\mu_\theta(s)$ and $\sigma_\theta(s)$ denote the state-dependent mean and standard deviation, respectively.

% For SAC and PPO type algorithms, the stochastic policy output the interaction actions follows:
% \begin{equation}
% \pi_\theta(a \mid s) = \mathcal{N}\!\big(\mu_\theta(s),\, \mathrm{diag}(\sigma_\theta^2(s))\big),
% \label{eq:sac_policy}
% \end{equation}

% 这里得解释为什么使用top O 
For efficient episodic memory retrieval, the current state $s_t$ is first encoded into a latent state embedding $z_{s_t}$. We then identify similar states in the failure episodic memory by measuring the $\ell_2$ distance in the embedding space with the threshold $\varepsilon$. Among the retrieved candidates, we further select the top $O$ state--action pairs associated with the lowest Monte Carlo returns, corresponding to the most hazardous past experiences. 

A scoring mechanism is introduced to evaluate candidate actions. For each candidate action, the score is composed of two terms: (1) the aggregated $\ell_2$ distance $D_i$ between its joint state–action embedding and the state–action embeddings retrieved from the memory; (2) the risk value $\rho_i$ of the corresponding state–action pair estimated by the trained risk head.
\begin{equation}
S_i = D_i - \lambda_{\text{risk}} \, \rho_i
\label{eq:risk_aware_score}
\end{equation}

%最后我们输出得分最高的候选动作作为最终参与交互的动作。
Finally, the highest-scoring candidate action is selected for interaction with the environment.

\section{Experiments}
We integrate FEMA with multiple model-free reinforcement learning algorithms and conduct a series of experimental validations on MuJoCo evaluation tasks and in a real-world bipedal robot stair-climbing task. The goals of the experimental evaluation are as follows: 1) Evaluate the effectiveness of FEMA when combined with different model-free RL methods and compare its performance against classical episodic control methods; 2) Assess the real-world applicability of FEMA; 3) Investigate the role of key hyperparameters and modules of FEMA through ablation studies.
%我们将FEMA与多种model-free强化学习算法结合，在贴近真实控制场景的MuJoCo评测任务中和真实双足机器人楼梯任务中进行了一系列实验验证。实验评估的目标如下：(1）在仿真实验中证明FEMA模块与任意model-free方法结合的有效性，以及相较于经典episodic control的优势；（2）结合PPO并行训练框架的真机测试验证FEMA的真机应用潜力；（3）消融实验来解释不同超参数的作用和选择以及方法框架的有效性。

\begin{figure*}[h]
\centering
\includegraphics[width=1.6\columnwidth]{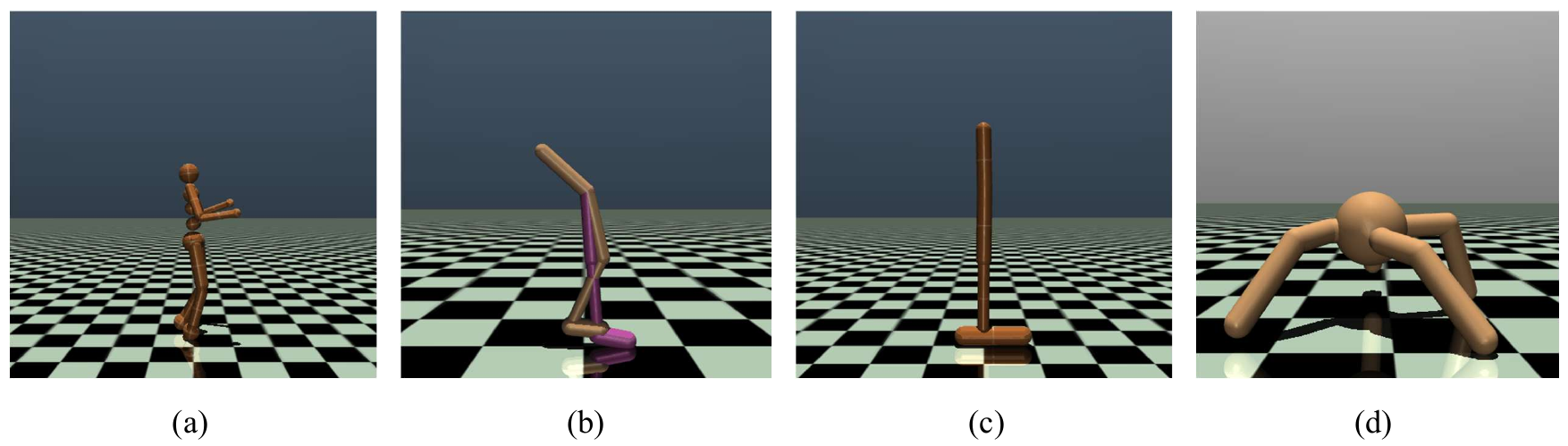}
\caption{MuJoCo evaluation tasks:(a)Humanoid (b)Walker2d (c)Hopper (d)Ant}
\label{figure:exp_settings}
\end{figure*}

\begin{figure*}[h]
\centering
\includegraphics[width=2.0\columnwidth]{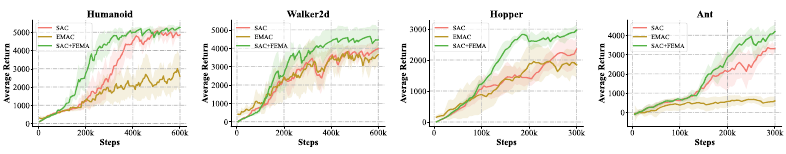}
\caption{Learning curves of EMAC, SAC, SAC+FEMA on MuJoCo benchmark tasks.}
\label{figure:s1}
\end{figure*}

\begin{figure*}[h]
\centering
\includegraphics[width=2.0\columnwidth]{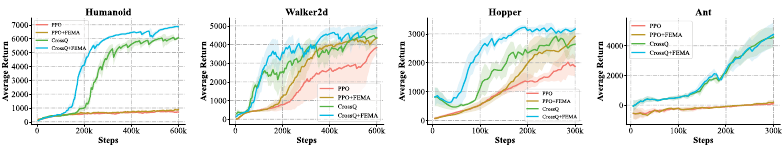}
\caption{Evaluation of FEMA integrated with PPO and CrossQ on MuJoCo benchmark tasks.}
\label{figure:s2}
\end{figure*}

\begin{table*}[t]
    \centering
    \caption{Maximum average return over four MuJoCo benchmark tasks across 5 random seeds.}
    \label{tab:mujoco_results}
    \resizebox{\linewidth}{!}{
    \begin{tabular}{lccccccc}
        \toprule
        \textbf{Task} &
        \textbf{SAC} &
        \textbf{EMAC} &
        \textbf{SAC+FEMA} &
        \textbf{PPO} &
        \textbf{PPO+FEMA} &
        \textbf{CrossQ} &
        \textbf{CrossQ+FEMA} \\
        \midrule
        Humanoid &
        $5290.80 \pm 127.64$ &
        $3808.84 \pm 1028.65$ &
        $5429.08 \pm 90.24$ &
        $902.71 \pm 190.19$ &
        $1102.82 \pm 599.42$ &
        $6267.30 \pm 251.07$ &
        $\mathbf{6969.74 \pm 69.13}$ \\
        Walker2d &
        $4102.53 \pm 339.45$ &
        $4538.35 \pm 283.47$ &
        $4728.77 \pm 505.40$ &
        $4004.54 \pm 217.06$ &
        $4538.86 \pm 175.93$ &
        $4580.71 \pm 514.70$ &
        $\mathbf{5026.85 \pm 250.32}$ \\
        Hopper &
        $3277.54 \pm 110.63$ &
        $2779.42 \pm 596.64$ &
        $3404.39 \pm 102.20$ &
        $2433.97 \pm 637.51$ &
        $3239.78 \pm 22.33$ &
        $3304.44 \pm 57.58$ &
        $\mathbf{3475.69 \pm 52.06}$ \\
        Ant &
        $4174.66 \pm 519.78$ &
        $832.98 \pm 20.04$ &
        $4757.64 \pm 405.30$ &
        $228.52 \pm 132.58$ &
        $427.70 \pm 395.18$ &
        $4875.93 \pm 732.88$ &
        $\mathbf{5082.74 \pm 841.68}$ \\
        \bottomrule
    \end{tabular}
    }
\label{table:max_reward}
\end{table*}

\subsection{Experiment Settings}

% 仿真实验中，我们这里选择了MuJoCo中的四个带termination标志的控制任务，包括Humanoid, Walker, Hopper, Ant。 当机器人的状态被判断危险时，往往对应即将摔倒或者发生碰撞，当前训练回合将会即刻中止。在真实实验中，我们将FEMA架构和当前主流的PPO并行训练架构进行融合，选择双足机器人楼梯任务作为测试任务。在仿真实验中，我们将FEMA模块和与PPO、SAC、CrossQ多种model-free算法进行结合，并于经典的episodic control方法EMAC进行对比。在不同的任务上，有不同的超参数设置，主要涉及候选动作的数量N，判断相似状态的阈值$\varepsilon$，和episodic memory的更新频率M。各个算法在不同任务上的超参数配置我放在了附录中，同时在4.3章节的消融实验中我也会讨论各个超参数的作用以及如何选择。
In the simulation experiments, we consider four tasks with explicit termination conditions from the MuJoCo benchmark, namely Humanoid, Walker2d, Hopper, and Ant, as shown in Figure~\ref{figure:exp_settings}. An episode is terminated immediately once the robot enters a hazardous state, such as an imminent fall or collision, which reflects safety constraints commonly encountered in real-world control scenarios. FEMA is combined with several representative model-free reinforcement learning algorithms, including PPO~\cite{schulman2017proximal}, SAC~\cite{haarnoja2018soft}, and CrossQ~\cite{bhatt2024crossq}, and is compared against the classical episodic control baseline EMAC~\cite{kuznetsov2021solving}.
All the above algorithms adopt the best-performing hyperparameter configurations as reported in their respective papers.
%上述的算法都使用他们论文中提及的最佳参数配置。
In the real-world experiments, we integrate the proposed FEMA technique into a widely used parallelized PPO training pipeline and validate it on a bipedal robot stair-climbing task.
The FEMA module involves several specific hyperparameters. These include the number of candidate actions $N$, which is typically set to 5 or 10, the state-matching similarity threshold 
$\varepsilon$, usually ranging from 0.03 to 0.10, and the episodic memory update frequency 
$M$, commonly set between 100 and 300. The optimal hyperparameter configurations vary across algorithms and tasks. Detailed settings for all experiments are provided in the appendix of the supplementary materials.
% Different tasks adopt different hyperparameter settings, primarily involving the number of candidate actions $N$, the similarity threshold $\varepsilon$ for state matching, and the update frequency $M$ of the episodic memory. Detailed hyperparameter settings for all algorithms and tasks are provided in the appendix of the supplementary materials.
Furthermore, in Section ~\ref{s4-4}, we conduct ablation studies to analyze the influence of these hyperparameters and discuss practical guidelines for their selection.

\begin{figure}[h]
\centering
\includegraphics[width=0.8\columnwidth]{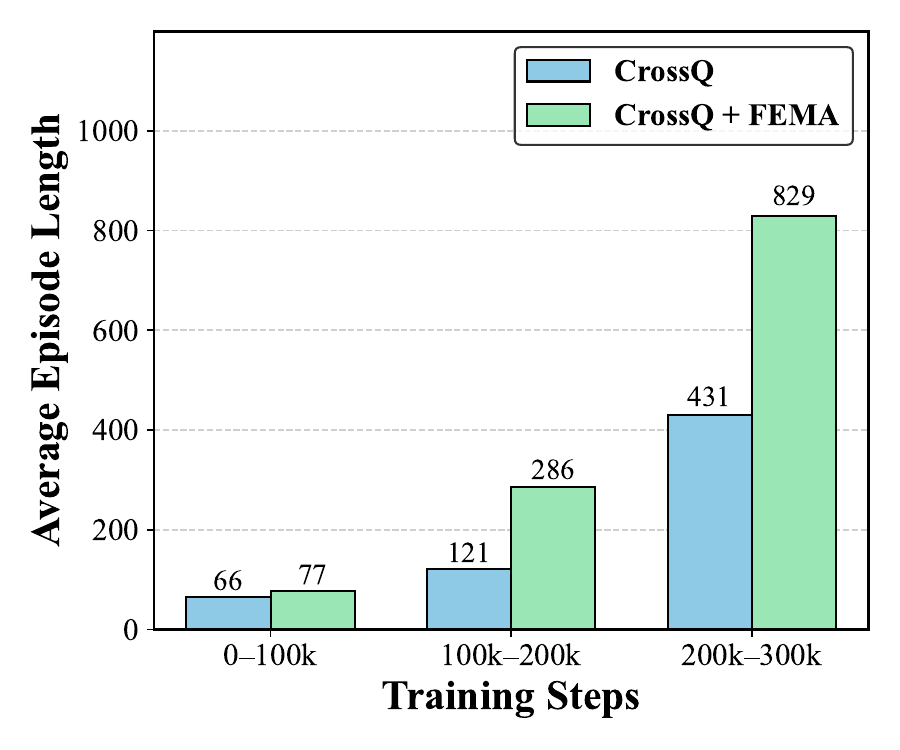}
\caption{Comparison of average episode length in the Humanoid task.}
\label{figure:episode_length}
\end{figure}

\begin{figure}[h]
\centering
\includegraphics[width=0.8\columnwidth]{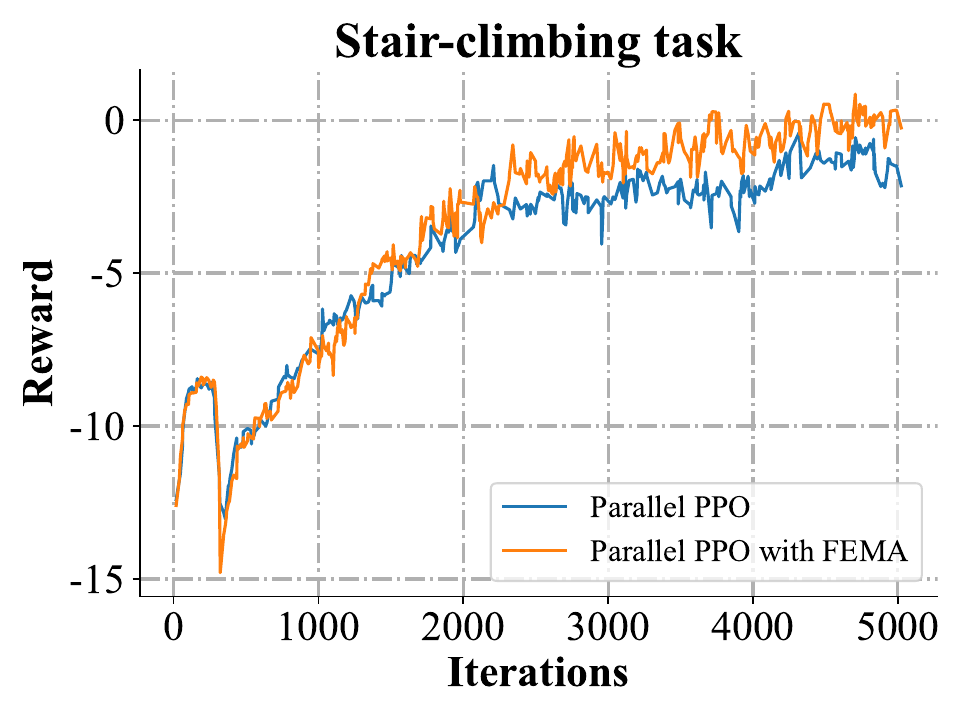}
\caption{Evaluation of FEMA integrated with parallel PPO in the real robot stair-climbing task.}
\label{figure:real_learning_curve}
\end{figure}

\begin{figure}[h]
\centering
\includegraphics[width=0.85\columnwidth]{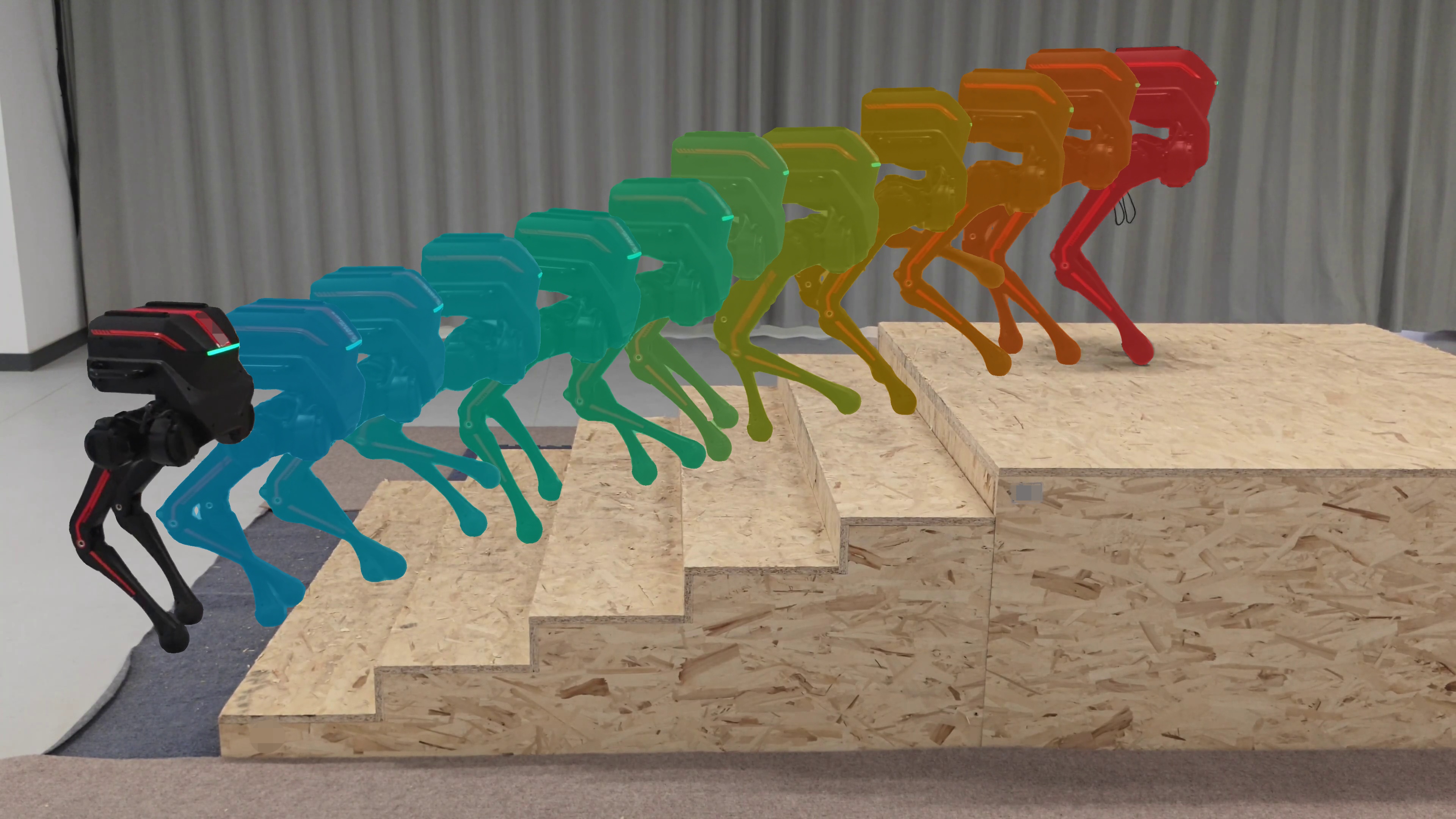}
\caption{Trajectory of the bipedal robot in the stair-climbing task.}
\label{figure:real_bipedal}
\end{figure}

\begin{figure}[h]
\centering
\includegraphics[width=0.8\columnwidth]{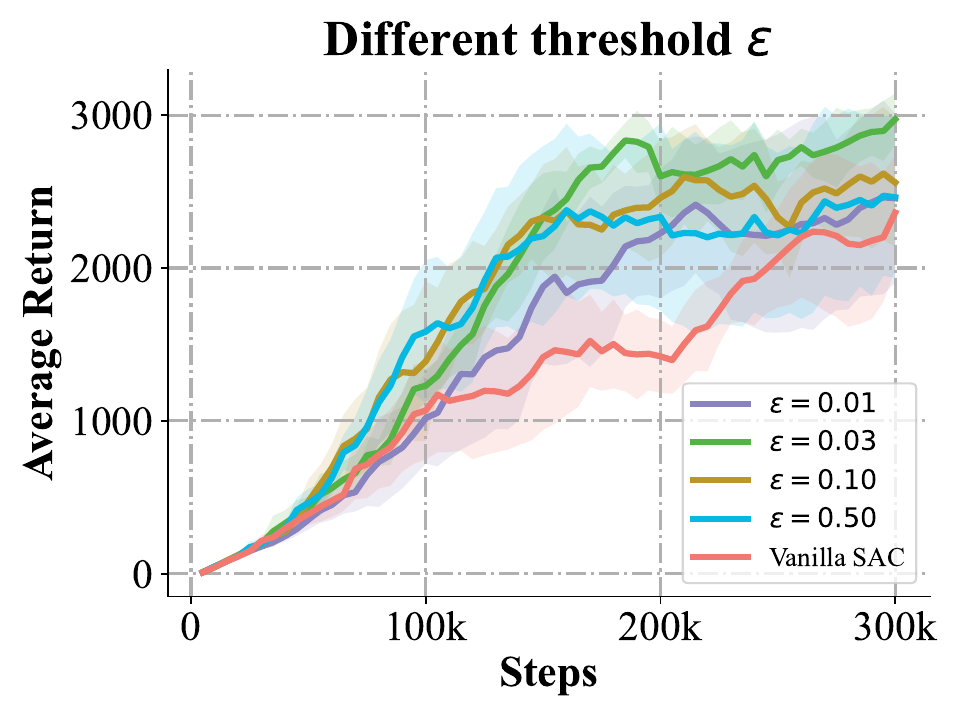}
\caption{Learning curves of SAC+FEMA with different $\varepsilon$.}
\label{figure:threshold}
\end{figure}

\begin{figure}[h]
\centering
\includegraphics[width=0.8\columnwidth]{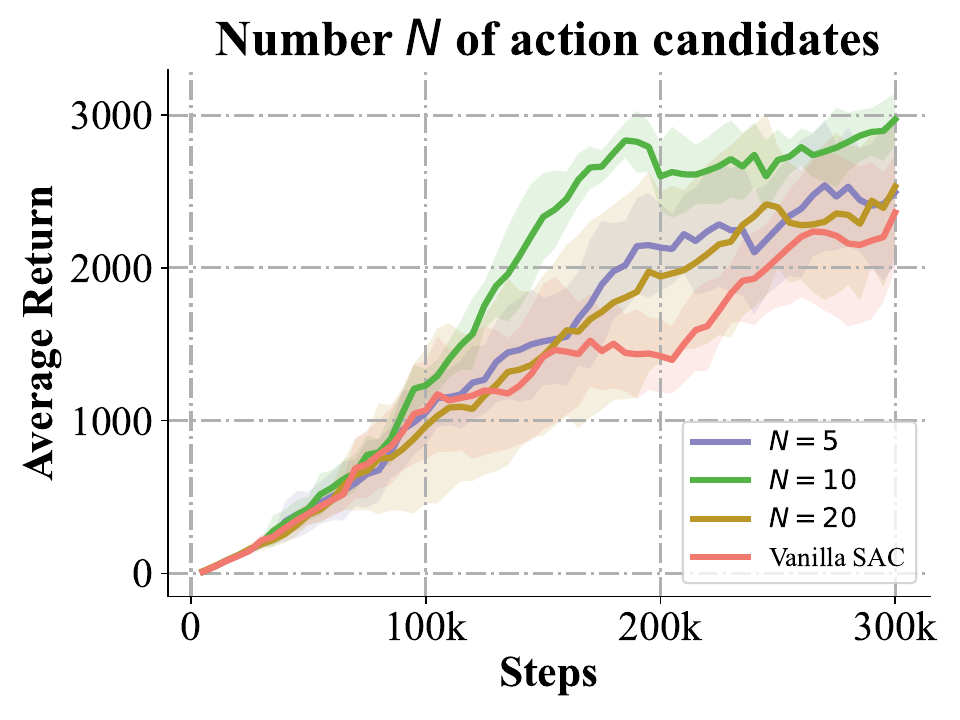}
\caption{Learning curves of SAC+FEMA with different candidate actions $N$.}
\label{figure:action_num}
\end{figure}

\begin{figure}[h]
\centering
\includegraphics[width=1.0\columnwidth]{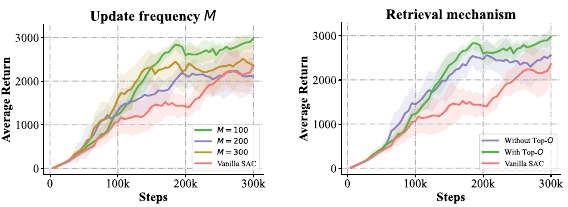}
\caption{Learning curves of SAC+FEMA under different episodic memory update frequency $M$ and ablation results of the Top-$O$ mechanism.}
\label{figure:frequency}
\end{figure}

% \begin{figure}[h]
% \centering
% \includegraphics[width=0.9\columnwidth]{ablation_notopk_learning_curve114.pdf}
% \caption{Trajectory of bipedal robot in the stair-climbing task}
% \label{figure:topk}
% \end{figure}

\subsection{Simulation Experiments}
%我们将FEMA与经典的model-free强化学习方法进行结合，并与同样基于actor-critic架构的EMAC方法进行对比，学习曲线如图4所示。
We first integrate FEMA with classical SAC algorithm to validate its efficacy. 
We also benchmark our method against EMAC, a representative episodic control baseline that also employs an actor-critic architecture. The learning curves are shown in Figure~\ref{figure:s1}. 
It is evident that integrating FEMA with SAC leads to significant improvements in both convergence speed and final performance across all four evaluation tasks. Moreover, as corroborated in Table ~\ref{table:max_reward}, the FEMA-augmented SAC method also attains a higher maximum average reward compared to the vanilla SAC baseline, indicating that the introduction of FEMA does not constrain the agent into sub-optimal, conservative behaviors.
%感觉最后少了一句
By comparison, EMAC performs poorly on all four evaluation tasks. These results suggest that conventional episodic control methods are ineffective when handling the abundance of short and failed episodes typically encountered during the early stages of learning. 
Using the maximum average return of vanilla SAC as the reference, SAC augmented with FEMA achieves 33.62\%, 61.86\%, 45.00\%, and 17.54\% improvements in sample efficiency on the Humanoid, Walker2d, Hopper, and Ant tasks, respectively.

We further integrate FEMA with PPO and the recent CrossQ algorithm to evaluate its general applicability, with the learning curves shown in Figure~\ref{figure:s2}. With FEMA, PPO achieves significantly faster convergence on the Walker2d and Hopper tasks, while on Humanoid and Ant, the enhanced long-horizon exploration enabled by FEMA leads to substantially higher maximum average returns compared to vanilla PPO. 
PPO equipped with FEMA consistently improves sample efficiency over vanilla PPO, yielding gains of 44.54\%, 25.45\%, 11.86\%, and 1.98\% on Walker2d, Hopper, Ant, and Humanoid, respectively.
Moreover, CrossQ+FEMA yields notable performance improvements on Humanoid, Walker2d, and Hopper. In contrast, the Ant task is relatively simple, where vanilla CrossQ can reach the maximum episode length easily through basic exploration and trial-and-error, resulting in limited additional gains from incorporating FEMA. Across the Humanoid, Walker2d, Hopper, and Ant tasks, CrossQ+FEMA demonstrates superior sample efficiency relative to vanilla CrossQ, with relative improvements of 48.74\%, 47.41\%, 50.98\%, and 8.33\%, respectively.

%我们还选择了Humanoid任务来深入探究FEMA技术对于训练进程的实际影响。如下图所示，我们统计了vanilla CrossQ和FEMA加持的CrossQ方法在训练中前的平均回合步长表现。不难看出，FEMA模块的引入使得智能体的前期训练收集得到更高长序列、高质量的数据，从而带来学习性能的提升。

The Humanoid task serves as a case study to examine the practical impact of FEMA on the training process. As illustrated in Figure~\ref{figure:episode_length}, we statistically analyzed the average episode length performance of the vanilla CrossQ algorithm and the FEMA-enhanced CrossQ variant during the early training phase.
It is evident that the integration of the FEMA module enables the agent to collect longer-horizon and higher-quality interaction data in the early training stage, thereby significantly boosting the overall learning performance of the algorithm. 
Between 200k and 300k training steps, the CrossQ method augmented with FEMA achieves an average episode length of 829 (with an upper limit of 1000), which is significantly higher than the 431 average episode length obtained by the vanilla CrossQ method.

% 在20万到30万steps训练阶段，可以看到有FEMA加持的CrossQ方法对应的回合平均步数达到829（上限1000），已经显著优于vanilla CrossQ方法对应的431平均轨迹长度。

% The results indicate that incorporating FEMA enables the agent to collect longer-horizon and higher-quality trajectories in the initial training phase, thereby leading to improved learning performance.

\subsection{Real-world Experiments}
We select a bipedal robot stair-climbing task as the real-world evaluation scenario. 
A bipedal robot with six degrees of freedom is required to climb stairs with a step height of 10 cm.
The robotic agent receives 30-dimensional proprioceptive observations, including base angular velocity, projected gravity, joint positions, joint velocities, previous actions, and gait phase features (sine/cosine clock signals and gait parameters).
The reward function combines velocity tracking and gait incentives (e.g., linear/angular velocity tracking and feet air-time) with strong stability and safety regularization.
Detailed training settings are provided in the appendix of the supplementary materials.
We adopt a widely used PPO-based parallel training framework in conjunction with the Isaac Gym platform to enable large-scale parallel learning. Specifically, 4,096 parallel environments are employed. 
As all robotic agents share interaction data under the PPO parallel training framework, the hyperparameter configurations of FEMA differ from those adopted in the MuJoCo benchmark tasks.
% As multiple robotic agents share interaction data during parallel training, the hyperparameters of FEMA is a little bit different. 
The episodic memory update interval $M$ is set to 2,500, meaning that the failure episodic memory is updated after every 2,500 collected failure trajectories. Apart from that, we set the threshold $\varepsilon$ to 0.20 and the number of candidate actions $N$ to 5.

% 因为多个robotic agents共享交互信息，这里我们将情景记忆的更新频率M设置为2500，即每收集到2500条失败轨迹则重新训练更新falure episodic memory中的信息。此外，阈值epison设置为0.20，候选动作N设置为5。
% As multiple robots share interaction data during parallel training, we increase the capacity of the episodic failure memory to store up to 2,500 failure trajectories. In addition, the episodic failure memory is updated in a batched manner after the collection of every 2,500 failure events. 

The learning curves are shown in Figure~\ref{figure:real_learning_curve}. It can be observed that, with the introduction of FEMA, parallel PPO converges to a higher reward within the same number of training steps.
The policy model obtained after 4,500 parallel training iterations is then deployed on a real robot and evaluated in a real stair-climbing environment. Representative trajectories of the test episodes are illustrated in Figure~\ref{figure:real_bipedal}, demonstrating that the learned policy is able to stably drive the bipedal robot to complete the stair-climbing task. 
Starting from the initial position, the bipedal robot reaches the top of the staircase in approximately 7 seconds.
By contrast, the vanilla parallel PPO algorithm fails to reliably complete the stair-climbing task even after 4,500 training iterations.
Its poor sim-to-sim transfer performance on the MuJoCo simulation platform leads us to exclude it from real-robot deployment and comparison.
% The bipedal robot reaches the top of the staircase from the initial position in approximately 7 seconds.
% The entire trajectory, from the initial state to the final state, takes approximately 7 seconds.
% 相较之下，vanilla parallel PPO算法在4500个iterations的训练之后无法稳定完成stair-climbing任务，sim2sim迁移至MuJoCo仿真平台上的糟糕表现让我们放弃了真机的部署和对照。
The performance improvements achieved by FEMA when integrated with parallel PPO further demonstrate its strong potential for real-world robot control applications.

\subsection{Ablation Study}\label{s4-4}
To enable the effective integration of FEMA with various RL algorithms, we investigate the selection of the three critical hyperparameters within the FEMA module: (1) the similarity threshold $\varepsilon$ used to retrieve relevant failure events; (2) the number $N$ of candidate actions sampled from the noisy policy distribution; (3) the memory update frequency $M$ of the failure episodic memory. We conduct ablation experiments on the Hopper task. For this task, FEMA-augmented SAC uses a similarity threshold of $\varepsilon=0.03$, the number of candidate actions $N=10$, and the memory update frequency $M=100$.

We first investigate the impact of different similarity thresholds $\varepsilon$ on learning performance. As illustrated in Figure~\ref{figure:threshold}, an excessively small threshold  $\varepsilon=0.01$ leads to substantially slower convergence in the early training stage, as the guidance provided by the Failure Episodic Memory becomes limited under strict similarity constraints. In contrast, an overly large threshold $\varepsilon=0.50$ causes FEMA to bias the policy toward conservative behaviors, leading to a substantially lower maximum return compared to the moderate setting of $\varepsilon=0.03$.

We further analyze the impact of the number of candidate actions on FEMA. Candidate actions are drawn in varying quantities from a Gaussian policy with added noise. As shown in Figure~\ref{figure:action_num}, the learning curves suggest that simply increasing the number of candidate actions does not guarantee improved performance. We also examine the impact of the Failure Episodic Memory update frequency $M$, as illustrated in the left part of figure~\ref{figure:frequency}. For the Hopper task, using a lower update frequency $M=200,300$ leads to degraded learning performance in the later training phase.

As described in Section~\ref{s3-3}, we introduce a strategy that incorporates only the top-$O$ matched events with the lowest Monte Carlo returns into candidate action scores ranking phase. We perform an ablation study on the Hopper task to assess this mechanism, as illustrated in the right part of Figure~\ref{figure:frequency}. The results suggest that the Top-$O$ strategy mitigates overly conservative behavior by focusing the guidance on the most critical failure experiences rather than aggregating all retrieved events.

\section{Conclusion}
We propose a novel technique termed Failure Episodic Memory Alert (FEMA), which enables robotic agents to achieve more efficient exploration during training. 
The failure episodic memory module effectively exploits the rich spatiotemporal correlations embedded in abundant short-horizon failure trajectories. Combined with the risk-aware action selection mechanism, the agent is guided to explore longer-horizon, higher-value experiences.

As a plug-and-play module, FEMA demonstrates consistent performance improvements when integrated with various model-free reinforcement learning algorithms on MuJoCo benchmark tasks. 
Furthermore, the integration of FEMA with the PPO-based parallel training framework and real-robot evaluation reveals its significant potential for real-robot control tasks.

\bibliographystyle{named}
\bibliography{ijcai26}

\end{document}